\documentclass{article}
\usepackage{spconf,amsmath,epsfig}
\usepackage{makecell}
\usepackage{booktabs}
\usepackage{float}

\title{HGDNet: A Height-Hierarchy Guided Dual-Decoder Network for Single View Building Extraction and Height Estimation}
\name{\parbox{\linewidth}{\centering Chaoran Lu$^{\dagger}$\thanks{ $^{\dagger}$ Equal contributions}, Ningning Cao$^{\dagger}$, Pan Zhang$^{*}$\thanks{ $^{*}$ Corresponding author, zhangpan@piesat.cn}, Ting Liu, Baochai Peng, \\ Guozhang Liu, Mengke Yuan, Sen Zhang, Simin Huang, Tao Wang}}
\address{PIESAT Information Technology Co, Ltd., Beijing, China}
\begin{document}
%
\maketitle
\begin{abstract}
Unifying the correlative single-view satellite image building extraction and height estimation tasks indicates a promising way to share representations and acquire generalist model for large-scale urban 3D reconstruction. However, the common spatial misalignment between building footprints and stereo-reconstructed nDSM height labels incurs degraded performance on both tasks. To address this issue, we propose a \textbf{H}eight-hierarchy \textbf{G}uided \textbf{D}ual-decoder \textbf{Net}work (HGDNet) to estimate building height. Under the guidance of synthesized discrete height-hierarchy nDSM, auxiliary height-hierarchical building extraction branch enhance the height estimation branch with implicit constraints, yielding an accuracy improvement of more than $6\%$ on the DFC 2023 track2 dataset. Additional two-stage cascade architecture is adopted to achieve more accurate building extraction. Experiments on the DFC 2023 Track 2 dataset shows the superiority of the proposed method in building height estimation ($\delta_1$:0.8012), instance extraction ($AP_{50}$:0.7730), and the final average score 0.7871 ranks in the first place in test phase.







\end{abstract}
\section{Introduction}
\label{sec:intro}

Extracting building footprints and heights from a single-view satellite image has arouse great interests in both academia and industry. With these attributes, a 3D city can be quickly reconstructed, making it an important technology for urban digital twins, urban management, etc. The two tasks have been thoroughly studied individually. Cascade Mask R-CNN~\cite{cai2019cascade}, and HTC++~\cite{liu2021swin} extract buildings with instance segmentation through multi-stage architectures. PopNet~\cite{zheng2019pop} and SCENet~\cite{xing2022sce} estimate building height with assisted semantic labels. Joint building extraction and height estimation are under investigated due to the scarcity of high quality geo-registered building footprints and nDSM labels. 

\begin{figure}[ht]

\begin{minipage}[b]{1.0\linewidth}
  \centering
\includegraphics[width=0.82\textwidth]{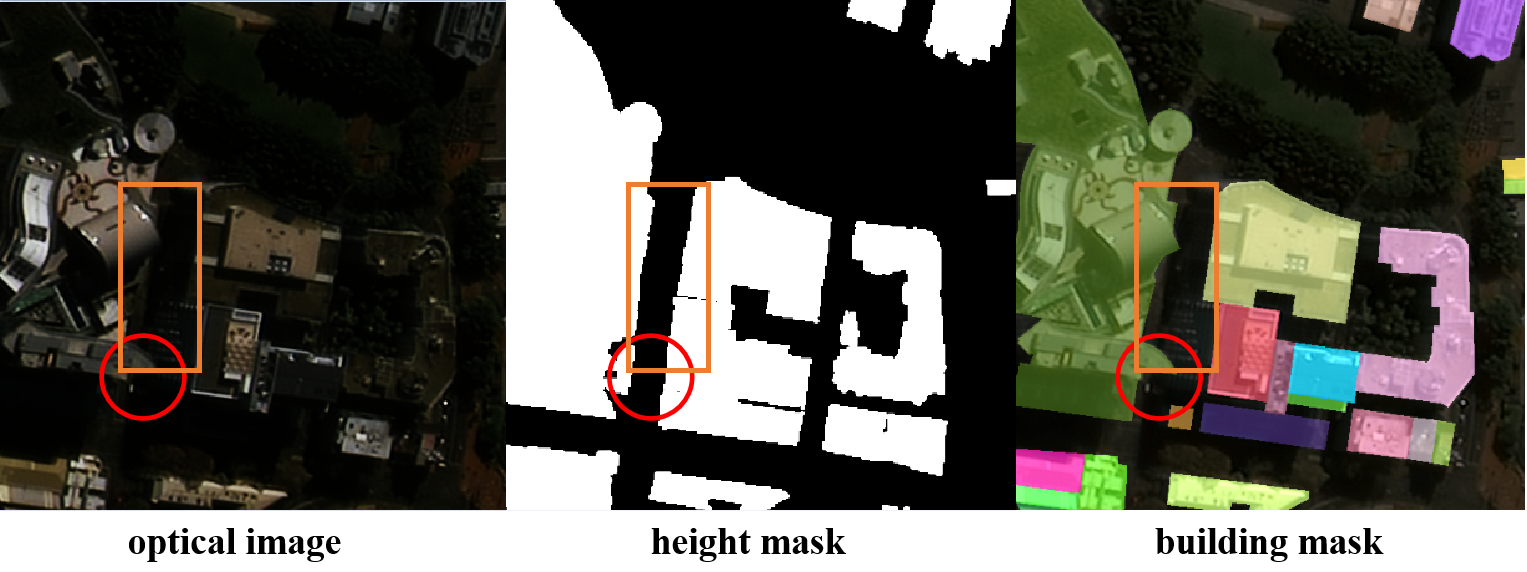}
\end{minipage}
\caption{A Misaligned Example in DFC2023 Dataset\cite{mrnt-8w27-22}.}
\label{fig:Misaligned}
\end{figure}

  
The IEEE GRSS DFC 2023 dataset~\cite{mrnt-8w27-22} provides a large scale building interpretation benchmark for building extraction and height estimation, which includes building instance segmentation labels and nDSMs. We have observed the misalignment between the two kind of labels as shown in Fig.~\ref{fig:Misaligned}, which probably caused by different data sources, as nDSMs are obtained by stereo reconstruction and have larger ground sampling distance. Meanwhile, the great quantity bias among high, medium and low elevation buildings brings difficulties in accurate height regression for all the height hierarchies.    





Therefore, We argue that the prerequisite of accuracy improvement of joint building extraction and height estimation should be achieved with reasonable building footprints and nDSMs supervision. Firstly, a novel dual-decoder height estimation model is proposed, the auxiliary height-hierarchical building extraction branch is integrated to alleviate the difficulty of directly height estimation for buildings of different heights. The nDSMs are pixel-wisely classified with discrete height hierarchies to generate newly synthesized building segmentation guidance map. Implicit constraints are formed between the height estimation and height-hierarchy building extraction branch. To fulfill precise building instances extraction, we employ a two-stage instance segmentation network to extract individual building contours. Extensive experiments and ablation studies on the DFC 2023 Track 2 dataset demonstrate the superiority of the proposed method in building height estimation ($\delta_1$:0.8012), instance extraction ($AP_{50}$:0.7730) and the final average score 0.7871 ranks in the first place in test phase.



\begin{figure*}[ht]
\centering
\includegraphics[width=0.75\linewidth]{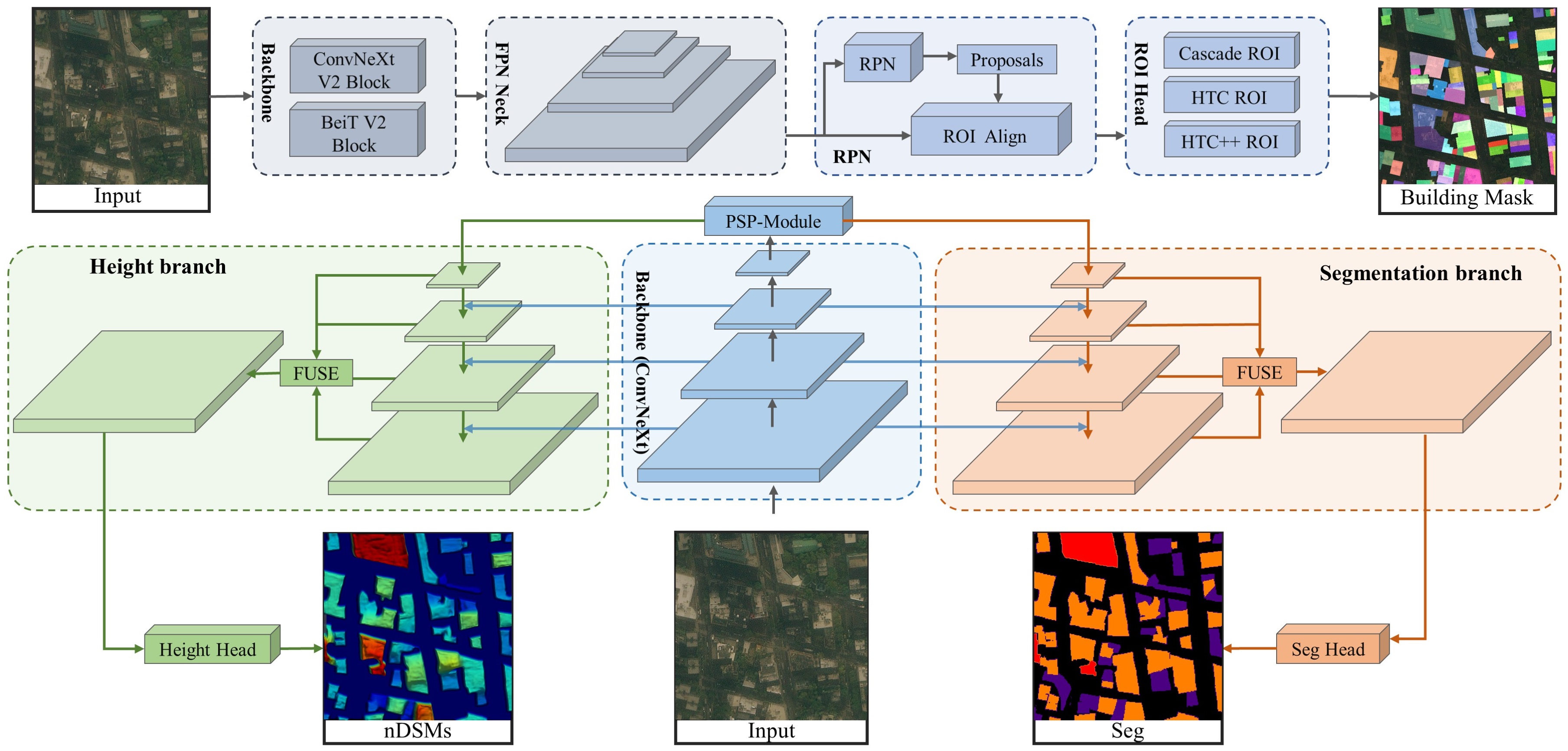}
\caption{The overview of our proposed building extraction(above) and height estimation(below) framework.}
\label{fig:network}
\end{figure*}

\section{Method}
In this section, firstly a novel dual-decoder network is presented, along with an explanation of how the height hierarchy is used to guide the regression of building heights. The instance segmentation models used for building extraction are then introduced.

\subsection{Height Estimation} 

The architecture of HGDNet is shown in Fig.\ref{fig:network}. ConvNeXt V2-Base is adopted as the encoder module. The dual decoders are consist of the height estimation branch and height-hierarchical segmentation branch.

\textbf{Height Estimation Branch:} 
Building heights are unevenly distributed, with a concentration between 0-50m, and even 0-10m.
Thus, the nDSMs is processed using a logarithmic function to make its distribution closer to normal. Then, maximum normalization is performed to facilitate faster model convergence. The normalized nDSMs is calculated as follows:
\begin{equation}
    DSM_{norm} = \frac{\ln{(nDSMs)}}{\max(\ln{(nDSMs)})}
\end{equation}

UperNet~\cite{xiao2018unified} is adopted as the basis for decoding layer. Multi-scale features from the backbone are enhanced via the up-down pathway and lateral connections. All enhanced features are upsampled to a certain scale and fused afterwards, which serves as the final feature for height estimation. Furthermore, an additional 1-channel convolutional layer and a sigmoid layer are added at the end of the height branch.
    
\textbf{Height-hierarchical Segmentation Branch:} 
To improve the accuracy of height estimation, an additional height-hierarchical segmentation branch is incorporated into the network. In this branch, the same UPerNet decoder as the height estimation branch is used. However, the fused features require only one $n$-channels convolutional layer, where $n$ is equal to the number of height hierarchies.
    
The proposed approach divides nDSMs into several discrete hierarchies rather than directly using instance segmentation annotations with only a single class. By analyzing the distribution of nDSMs and using clustering algorithms, nDSMs is divided into $n$ classes as height-hierarchy labels. And these discrete labels are used to guide the estimation of the building heights.

\textbf{The Weighted Loss Function:}
The height-hierarchical segmentation branch uses Cross-Entropy loss, and the height branch uses $smooth_{L1}$ loss. Different loss weights are applied in two branches. The finally loss function is as below:

\begin{equation}
    Loss = \alpha * L_{CE} + \beta * L_{smooth_{L1}}
\end{equation}

\subsection{Building Extraction}
Multiple two-stage cascade detection frameworks and backbones are adopted in building extraction task.

\textbf{Cascade Framework:} Several efficient cascade networks are adopted such as Cascade Mask R-CNN~\cite{cai2019cascade}, HTC++\cite{liu2021swin}. These networks extract targets in two stages, first focus on detecting the location of the targets to obtain proposals, and then classify and segment while correct the proposals.

\textbf{Backbone:} After comparative experiments, ConvNeXt V2-Base\cite{Woo2023ConvNeXtV2} and BEiT V2\cite{peng2022beit} networks are chosen for extracting semantic features. These backbone networks extract deep features at different scales from inputs to satisfy decent extraction results of buildings at different scales.

\textbf{Loss Function:}  $smooth_{L1}$ loss function and GIOU loss function are used to make the bounding box more closely to the detected buildings, and the Cross-Entropy loss function is used in building mask learning.

\textbf{Model Ensemble:} We modified Weighted Boxes Fusion (WBF\cite{solovyev2021weighted}) as Weighted Segmentation Fusion (WSF): first, adopt WBF to obtain fused bboxes, then fuse the segment masks, finally crop the fused masks by fused boxes. This strategy is used for integrating different models with multi-scale. 

\section{EXPERIMENTS}
\label{sec:experiment}

\subsection{Dataset}
\label{ssec:dataset}

The DFC2023 dataset~\cite{mrnt-8w27-22} is based on Urban Building Classification (UBC)\cite{huang2022urban}, and is supplemented with SAR imagery and nDSMs at the same location. The whole dataset contains 2957 tiles with $512\times512$ pixels that are resampled to 0.5m. But as observation like Fig.\ref{fig:Misaligned}, the optical image, nDSMs-based mask, and building instance annotation in the dataset are not aligned with each other. Besides, the final solution does not utilize SAR images because SAR and optical images are not strictly aligned, and experiments also indicate that SAR images bring no improvement in this challenge.

\subsection{Implementation Details}
\quad\ \textbf{Height Estimation:}
In height estimation task, we set the batch size as 2, the initial learning rate as 1e-4, and conduct 16k iterations of training, during which the learning rate is decreased using the polynomial decay with power 1.0. The parameters of the AdamW optimizer are aligned with the building extraction task. In the additional height-hierarchical segmentation branch, buildings are classified into 4 categories based on heights, i.e., ground-wise [0, 1e-6), low-wise [1e-6, 10), medium-wise [10, 36) and high-wise [36, 187). The training data is resized randomly to 0.5-2.0 times of the original image, and random rotations are used as data augmentation. We incorporate the height estimation outcomes using the max function in conjunction with multi-scale inference. The loss function coefficients of height-hierarchical segmentation branch and height branch are 5 and 30, respectively.

\textbf{Building Extraction:}
All models are trained with a batch size of 2. We use 36 epochs for training, with initial learning rate of 0.0004 and decaying by a factor of 0.1 at the 27th and 33rd epoch. The AdamW optimizer descent with a weight decay of 0.05 and $\beta_1=0.9$, $\beta_2=0.999$. To accommodate buildings with different scales, the anchor scales of Region Proposal Network is set to 4 and 8. During inference stage, test data are resized to 6-9 scales between 400 and 1400, and then these multi-scale results are fed to WSF processing.

\textbf{Corrective Process of Heights:}
The outcomes of the height-hierarchical segmentation branch are considered to correct the height estimates. Generally, building heights are not less than 3m, so the height estimates are corrected to 0 at locations where the prediction value in height-hierarchical segmentation branch is 0 while the prediction value in height branch is below 3m.

\subsection{Main Results}
\begin{figure}[ht]
\centering
\includegraphics[width=0.45\textwidth]{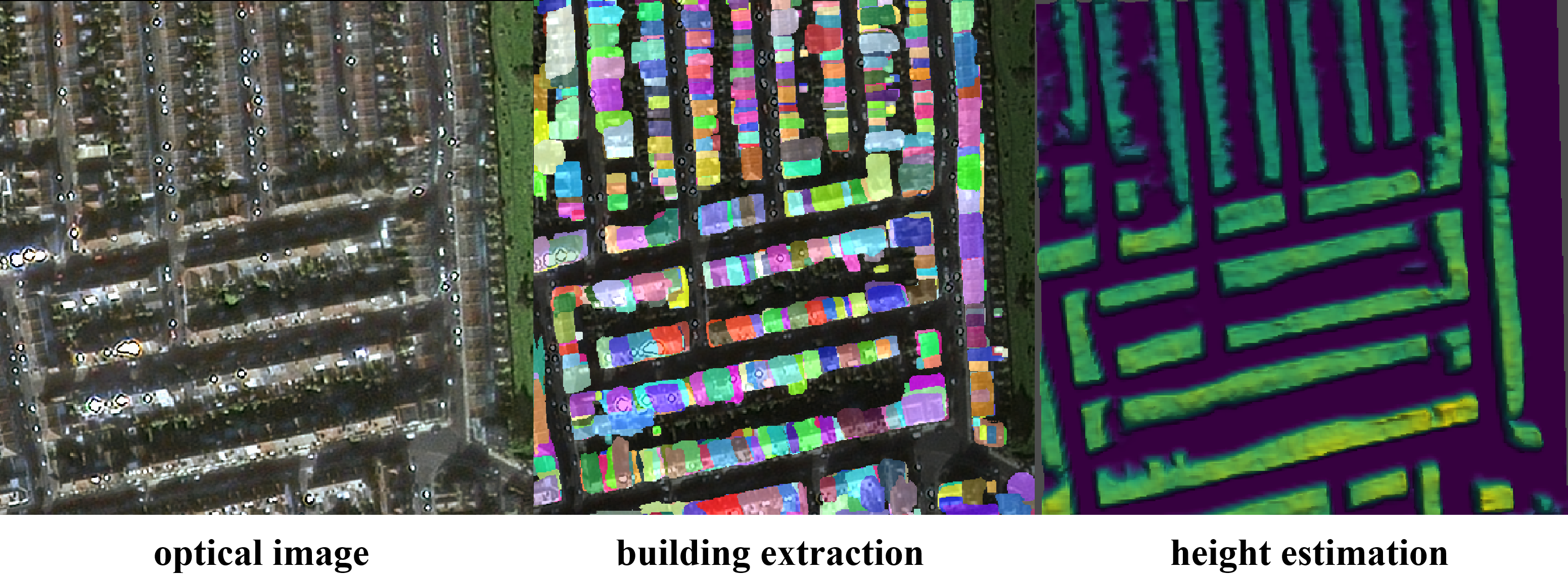}
\caption{Experiment Results}
\label{fig:results}
\end{figure}

The proposed solution has achieved a performance of 0.773 for the $AP_{50}$ of building extraction and 0.8012 for the $\delta_1$ of height estimation on the test set. The total score is 0.7871, ranking 1st in the multi-task learning of joint building extraction and height estimation challenge of the 2023 IEEE GRSS Data Fusion Contest. One example is shown in Fig.~\ref{fig:results}.

 The accuracy metrics of each model are shown in Tab.\ref{tab:segresults}. The best performing result is achieved with seven models, which were not the top seven highest-scoring models, but rather seven models with different architectures.

\begin{table}[htb]
    \caption{Metrics for 7 building extraction models and their fusion result on the test set.}
    \centering
    \begin{tabular}{c c c}
    \toprule
        Method & $AP_{50}$ & $mAP$ \\
    \bottomrule
        HTC++\_BEiT V2 & 0.730 & 0.402 \\
        HTC++\_ConvNeXt V2-Base & 0.710 & 0.388 \\
        HTC\_BEiT V2 & 0.722 & 0.400 \\
        \makecell[c]{Cascade\_BEiT V2} & 0.751 & 0.428 \\
        \makecell[c]{Cascade\_ConvNeXt V2-Base} & 0.737 & 0.422 \\
        HTC\_ConvNeXt V2-Base & 0.732 & 0.420 \\
        \makecell[c]{Cascade\_ ConvNeXt V2-Base } & 0.721 & 0.410 \\
    \toprule
        \textbf{WSF based model ensemble } & \textbf{0.773} & \textbf{0.450} \\
    \bottomrule
    \end{tabular}

    \label{tab:segresults}
\end{table}

The performance of our height estimation model are compared with other methods. Tab.\ref{tab:different height methods} shows the specific quantitative comparison.

\begin{table}[htb]
    \caption{Comparison of our method and other three methods on validation set. The last line is metrics of our method on test set. *Baseline is officially provided by the organization.}
    \centering
    \begin{tabular}{c c c c}
    \toprule
        Method & $\delta_1$ & $\delta_2$ & $\delta_3$ \\
    \bottomrule
        $Baseline^*$ & 0.7042 & 0.7613 & 0.7922 \\
        DSMNet\cite{elhousni2021height} & 0.7339 & 0.7866 & 0.8110 \\
        SCENet-50\cite{xing2022sce} & 0.7731 & 0.8214 & 0.8438 \\
        Ours (validatation set) & \textbf{0.7966} & \textbf{0.8383} & \textbf{0.8581} \\
    \toprule
        Ours (test set) & 0.8012 & 0.8460 & 0.8674\\
    \bottomrule
        
    \end{tabular}
    \label{tab:different height methods}
\end{table}

\subsection{Ablation Study}
 A detailed comparison is shown below  to investigate the functionality of backbone and height-hierarchical segmentation branch in the proposed building height estimation method. 
 
\textbf{Height-hierarchical Segmentation Branch:}
The height estimates are corrected using the height-hierarchical segmentation branch results. Tab.\ref{tab:with seg or not} shows the impact of the segmentation branch to the model. Obviously, height estimation can benefit from semantic segmentation.

\begin{table}[htb]
 \caption{Quantitative comparison of whether to add height-hierarchical segmentation branch on validation set.}
    \centering
    \begin{tabular}{c c c c}
    \toprule
        Method & $\delta_1$ & $\delta_2$ & $\delta_3$ \\
    \bottomrule
        Without seg branch & 0.7313 & 0.7822 & 0.8084 \\
        With seg branch & \textbf{0.7966} & \textbf{0.8383} & \textbf{0.8581} \\
    \bottomrule
    \end{tabular}
    \label{tab:with seg or not}
\end{table}

\textbf{Backbone:}
The backbone is used to extract features of input images and is crucial to the model. We compare the accuracy metrics of three backbone networks in Tab.\ref{tab:different backbone}.

\begin{table}[htb]
    \caption{Quantitative comparison of different backbones.}
    \centering
    \begin{tabular}{c c c c}
    \toprule
        Backbone & $\delta_1$ & $\delta_2$ & $\delta_3$ \\
    \bottomrule
        ResNet-50 & 0.7824 & 0.8037 & 0.8502 \\
        ConvNeXt V1 Base & 0.7930 & 0.8371 & 0.8544 \\
        ConvNeXt V2 Base & \textbf{0.7966} & \textbf{0.8383} & \textbf{0.8581} \\
    \bottomrule
    \end{tabular}
    \label{tab:different backbone}
\end{table}

\section{CONCLUSION}

In this paper, we have addressed the problems of instance-level buildings extraction and building height estimation in single-view remote sensing imagery. A simple but effective Height-hierarchy Guided Dual decoders architecture(HGDNet) is proposed to estimate building heights.  Semantic segmentation task and height estimation task share multi-scale features of an advanced backbone, and form implicit constraint to achieve better performance. Additionally, the results of multiple two-stage instance segmentation models are fused with multi-scale, ensuring that the proposed model can recognize buildings of different scales. Experimental results on DFC2023 dataset demonstrates the effectiveness of the proposed method.

\section{ACKNOWLEDGEMENT}
The authors would like to thank the IEEE GRSS Image Analysis and Data Fusion Technical Committee, Aerospace Information Research Institute, Chinese Academy of Sciences, Universität der Bundeswehr München, and GEOVIS Earth Technology Co., Ltd. for organizing the Data Fusion Contest. This work was supported in part by the Special Funds for Creative Research (No. 2022C61540).


\bibliographystyle{IEEEbib}
\bibliography{refs2}

\end{document}